\documentclass[11pt,a4paper]{article}
\usepackage[hyperref]{acl2021}
\usepackage{times}
\usepackage[utf8]{inputenc}

\usepackage{cprotect}
\usepackage{todonotes}
\usepackage{url}
\usepackage{indentfirst}
\usepackage{multirow}
\usepackage{comment}
\usepackage{xcolor}
\usepackage{xspace}
\usepackage{amssymb}
\usepackage{url}
\usepackage{float}
\usepackage{amsmath}
\usepackage[titlenumbered,ruled]{algorithm2e}
\usepackage{graphicx}
\usepackage{latexsym}
\usepackage{todonotes}
\usepackage{enumitem}

\usepackage{}

\usepackage{microtype}

\newcommand{\ours}{\textsc{ABCD}\xspace}

\aclfinalcopy 
\def\aclpaperid{2825} 

\title{\textbf{ABCD}: 
A Graph Framework to Convert Complex
Sentences \\
to a Covering Set of Simple Sentences
 }

\author{Yanjun Gao \and Ting-Hao (Kenneth) Huang \and Rebecca J. Passonneau \\
        Pennsylvania State University \\ 
        \tt { \{yug125,txh710,rjp49\}@psu.edu}}

\date{}

\makeatletter
\def\thickhline{%
  \noalign{\ifnum0=`}\fi\hrule \@height \thickarrayrulewidth \futurelet
   \reserved@a\@xthickhline}
\def\@xthickhline{\ifx\reserved@a\thickhline
               \vskip\doublerulesep
               \vskip-\thickarrayrulewidth
             \fi
      \ifnum0=`{\fi}}
\makeatother

\newlength{\thickarrayrulewidth}
\begin{document}
\maketitle

\begin{abstract}


Atomic clauses are fundamental text units for understanding complex sentences.
Identifying the atomic sentences 
within complex sentences 
is important for applications such as summarization, argument mining, discourse analysis, discourse parsing, 
and question answering.
Previous work mainly relies on rule-based methods dependent on parsing.
We propose a new task 
to decompose each complex sentence into
simple sentences derived from the tensed clauses in 
the source,
and a novel problem formulation 
as a graph edit task.
Our neural model learns to \textbf{A}ccept, \textbf{B}reak, \textbf{C}opy or \textbf{D}rop 
elements of a graph that combines word adjacency and grammatical dependencies.
The full
processing pipeline 
includes modules for graph construction, 
graph editing, and sentence generation from the output graph.
We introduce DeSSE, a new dataset designed to train and evaluate complex sentence decomposition, and MinWiki, a subset of 
MinWikiSplit.
\ours achieves comparable performance as two parsing baselines on MinWiki.
On DeSSE, which has a more even balance of complex sentence types, our model achieves higher accuracy on the number of atomic sentences than an encoder-decoder baseline.
Results include a detailed error analysis.
\end{abstract}

\section{Introduction}

Atomic clauses are fundamental text units for understanding complex sentences. The ability to decompose complex sentences 
facilitates research that 
aims
to identify, rank or relate 
\textit{distinct predications}, such as content selection in summarization~\cite{fang2016proposition, peyrard2017supervised}, 
labeling argumentative discourse units in
argument mining~\cite{jo-etal-2019-cascade} or elementary discourse units in
discourse analysis~\cite{mann1986relational,burstein2001enriching, demir2010discourse}, 
or extracting atomic 
propositions for question answering~\cite{pyatkin2020qadiscourse}.
In this work, we propose a new task to decompose 
complex sentences into a covering set of simple sentences, with one simple output sentence per tensed clause in the source sentence. 
We focus on tensed clauses rather than other constituents because they are syntactically and semantically more prominent, thus more essential in downstream tasks like argument mining, summarization, and question answering.

\begin{figure}[t]
 \small 
    \centering
    \begin{tabular}{l|l}
         Orig& \underline{Sokuhi was born in Fujian} \textcolor{red}{and} \underline{was ordained at 17}. \\
         SS1& Sokuhi was born in Fujian. \\ 
         SS2& \textcolor{blue}{Sokuhi}  was ordained at 17. 
    \end{tabular}
    \vspace{-.14in}
    \caption{\small Example of a complex sentence (Orig) 
    rewritten as two simple sentences (SS1, SS2). 
    Underlined words in the source are preserved in the same order in the two outputs, the conjunction \textit{and} (red font) is dropped,
    and the subject \textit{Sokuhi} (blue font) is copied to the second simple sentence. 
    }
    \label{fig:exampleCS}
\end{figure}


The complex sentence decomposition task we address has some overlap with related NLP algorithms, 
but each falls short in one or more respects.
Elementary discourse unit (EDU) segmentation 
segments source sentences 
into a sequence of
non-overlapping spans~\cite{carlson2003building, wang2018toward}. The output EDUs, however, 
are not always complete clauses. 
Text simplification 
rewrites complex sentences using simpler vocabulary and syntax~\cite{zhang2017sentence}.
The output, however, does not preserve every tensed clause in the original sentence. The split-and-rephrase (SPRP) task aims to rewrite complex sentences into sets of shorter sentences, 
where an output sentence can be derived from non-clausal constituents in the source~\cite{narayan2017split}. 
In contrast to the preceding methods, we convert each tensed clause in a source sentence, including each conjunct in a conjoined VP, into an independent simple sentence. Unlike EDU segmentation, a belief verb and its \textit{that}-complement do not lead to two output units. Unlike text simplification, no propositions in the source are omitted from the output. Unlike SPRP, a phrase that lacks a tensed verb in the source cannot lead to a distinct sentence in the output. 

Figure~\ref{fig:exampleCS} shows an example complex sentence (Orig) with conjoined verb phrases 
and its rewrite into two simple sentences (SSs).
Observe that besides producing two sentences from one, thus breaking the adjacency between words,
words inside the verb phrases (underlined in the figure) remain 
in the same linear order in the output; 
the single subject \textit{Sokuhi} in the source 
is copied to the more distant verb phrase. 
Finally, the connective \textit{and}  
is dropped. We find that most rewrites of complex sentences into simple sentences that preserve the one-to-one mapping of source tensed predication with target simple sentence involve similar operations. 
Building on these observations,
we propose a neural model that learns to \textbf{A}ccept, \textbf{B}reak, \textbf{C}opy or \textbf{D}rop 
elements of a special-purpose sentence graph that represents word adjacency and grammatical dependencies, 
so the model can learn based on both kinds of graph proximity.
We also introduce DeSSE (\textit{\textbf{De}composed \textbf{S}entences from \textbf{S}tudents \textbf{E}ssays}), a new annotated dataset to support our task. 

The rest of the paper presents two evaluation datasets, our full pipeline, and our \ours model. 
Experimental results show that \ours 
achieves comparable or better performance than baselines.  \footnote{\ours is available at \url{https://github.com/serenayj/ABCD-ACL2021}.}

\section{Related Work}


Related work
falls largely into parsing-based methods, neural models that rewrite, and neural segmenters.  
\citet{gao-etal-2019-automated} propose a decomposition parser (DCP) that extracts VP constituents and clauses from complex sentences as part of a summarization evaluation tool. \citet{niklaus-etal-2019-dissim} present a system (DisSim) 
based on parsing
to extract simple sentences from complex ones. 
\citet{jo2020machine} propose 
seven rules to extract 
complete propositions from parses of complex questions and imperatives for argumentation mining. 
Though performance of these methods depends on parser quality, they often achieve very good performance. We include two whose code is available (DCP, DisSim) among our baselines.

SPRP models are based on encoder-decoder architectures, and the output is highly depending on the training corpus.
\citet{aharoni2018split} present a Copy-augmented network (Copy\textsubscript{512}) based on~\cite{gu2016incorporating} that 
encourages the model to copy most words from the original sentence to the output. As it achieves improvement over an earlier encoder-decoder SPRP model 
\cite{narayan2017split}, 
we include Copy\textsubscript{512} among our baselines.

Finally, recent neural EDU segmenters~\cite{wang2018toward, li2018segbot} achieve state-of-the-art performance on a discourse relation corpus, RST-DT~\cite{carlson2003building}. 
As they do not output complete sentences, we do not include any among our baselines.

Our ABCD model leverages the detailed information captured by parsing methods, and the powerful representation learning of neural models. As part of a larger pipeline that converts input sentences to graphs, ABCD learns to predict graph edits for a post processor to execute.

\section{Datasets}

Here we present DeSSE, a corpus we collected for our task,
and MinWiki, 
a modification of 
an existing SPRP corpus (MinWikiSplit~\cite{niklaus-etal-2019-minwikisplit}) to support our aims. We also give 
a brief description of 
differences in their distributions. Neural models are heavily biased by the distributions in their training data~\cite{niven2019probing}, and we show that DeSSE has a more even balance of linguistic phenomena. 

\subsection{DeSSE}
\label{subsec:desse}
DeSSE is collected in an undergraduate social science class, where students watched video clips about race relations, and wrote essays in a blog environment to share their opinions with the class. It was created to support analysis of student writing, so that different kinds of feedback mechanisms can be developed regarding sentence organization. 
Students have difficulty with revision to address lack of clarity in their writing \cite{kuhn2016tracing}, such as non-specific uses of connectives, run on sentences, repetitive statements and the like. These make DeSSE different from corpus with expert written text, such as Wikipedia and newspaper.  
The annotation process is unique in that it involves identifying where to split a source complex sentence into distinct clauses, and how to rephrase each resulting segment as a semantically complete simple sentence, omitting any discourse connectives. It differs from corpora that identify discourse units within sentences, such as RST-DT~\cite{carlson2003building} and PTDB~\cite{prasad2008penn}, because clauses are explicitly rewritten as simple sentences. It differs from split-and-rephrase corpora such as MinWikiSplit, because of the focus in DeSSE on rephrased simple sentences that have a one-to-one correspondence to tensed clauses in the original complex sentence. DeSSE is also used for connective prediction tasks, as in~\cite{gao-etal-2021-learning}.\footnote{DeSSE and MinWiki 
are available at \url{https://github.com/serenayj/DeSSE}.}


\begin{figure}
\small
    \centering
   \begin{itemize}
    \setlength{\itemsep}{-1.8pt}
        \item \textbf{Orig}: (I believe that \textit{talking about race more in a civil way can only improve our society}), $\vert \vert$ {\bf but} I can see why other people may have a different opinion.
        \item \textbf{Rephrase 1}: I believe that talking about race more in a civil way can only improve our society. 
        \item \textbf{Rephrase 2}: I can see why other people may have a different opinion. 
    \vspace*{-.24in}
    \end{itemize}
    \caption{\small An original sentences from DeSSE with an intra-sentential connective (\textbf{but}), a verb that takes a clausal argument. The annotation first splits the sentence (at $\vert \vert$), then rephrases each segment into a simple sentence, dropping the connective.}
    \label{fig:complexS}
\end{figure}



We perform our task on Amazon Mechanical Turk (AMT). In a series of pilot tasks on AMT, we iteratively designed annotation instructions and an annotation interface, while monitoring quality. Figure~\ref{fig:complexS} illustrates two steps in the annotation: identification of $n$ split points between tensed clauses, and rephrasing the source into $n+1$ simple clauses, where any connectives are dropped. The instructions 
ask annotators to focus on
tensed clauses occurring in conjoined or subordinate structures, relative clauses, parentheticals, and conjoined verb phrases, and to exclude 
gerundive phrases, infintival clauses, and clausal arguments of verbs.
The final version of the instructions describes the two annotation steps, provides a list of connectives, and illustrates a positive and negative example.\footnote{In step 2, the interface checked for any remaining connectives, 
to warn annotators. Details about the interface and quality control are included in appendix~\ref{sec:appendixB}.} 
The training and tests sets contains 12K and 790 examples, respectively. 

\begin{table}[t!]
\begin{center}
\small
\begin{tabular}{l|r|r|r|r|r} \hline 
\multirow{ 2}{*}{Dataset} & Disc. & VP- & Wh- \& & Restric. & {\it that}- \\  
                         & Conn. & Conj. & Rel. Cl. & Rel. Cl. &  comp. \\\hline
MinWiki & 58\% & 36\% & 10\% & 26\% & 0\%  \\ 
DeSSE  & 66\% & 22\% & 32\% & 34\% & 24\% \\\hline 
\end{tabular}
\vspace{-.1in}
\caption{
\small Prevalence of five 
linguistic phenomena in 50 randomly selected 
examples per dataset. Categories are not mutually exclusive. }
\label{tab:lings}
\end{center}
\end{table}

 
\subsection{MinWikiSplit}
 
MinWikiSplit 
has \textsc{203K} complex sentences and their rephrased versions~\cite{niklaus-etal-2019-minwikisplit}. It is built from \textit{WikiSplit}, a text simplification dataset 
derived from Wikipedia revision histories
~\cite{narayan2017split}, modified to focus on 
minimal propositions that cannot be further decomposed. 
It was 
designed for simplifying complex sentences into multiple simple sentences, 
where the simple sentences can correspond to a very wide range of structures from the source sentences, such as 
prepositional or adjectival phrases. 
To best utilize this corpus for our purposes, we 
selected a subsample 
where the number of tensed verb phrases in the source sentences matches the number of rephrased propositions. 
The resulting MinWiki corpus has an 18K/1,075 train/test split. 
 
\subsection{Linguistic phenomena} 

Table~\ref{tab:lings} presents prevalence of 
syntactic patterns characterizing complex sentences 
in the two datasets. Four are 
positive examples of one-to-one correspondence of tensed clauses in the source with simple sentences in the rephrasings:
discourse connectives (Disc. Conn.), VP-conjunction, clauses introduced by {\it wh-} subordinating conjunctions (e.g., {\it when, whether, how}) combined with non-restrictive relative clauses ({\it wh-} \& Rel. Cl.), and restrictive relative clauses (Restric. Rel. Cl.). The sixth column (negative examples) covers clausal arguments, which are often {\it that-}complements of verbs that express belief, speaking, attitude, emotion, and so on. MinWiki has 
few of the latter, presumably due to the genre difference between opinion essays (DeSSE) and Wikipedia (MinWiki). 

\section{Problem Formulation}

We formulate the problem of converting complex sentences into covering sets of simple sentences as a graph segmentation problem.
Each sentence is represented as a \textit{Word Relation Graph} (WRG), a directed graph 
constructed 
from each input sentence with its dependency parse. 
Every word token and its positional index becomes a WRG vertex.
For every pair of words, one or more edges are 
added as follows: 
a \textit{neighbor} edge that indicates that the pair of words are linearly adjacent; a \textit{dependency} edge that shows every pair of words connected by a dependency relation, 
adding critical grammatical relations, such as \texttt{subject}.

\begin{figure}
 \centering
 \small
\includegraphics[width=\columnwidth]{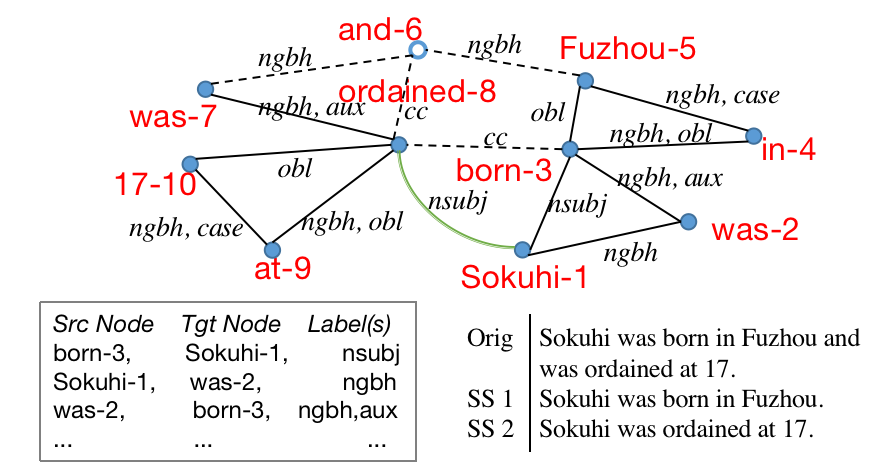}
\vspace{-.26in}
\caption{\small Example complex sentence (Orig), ground truth output (SS 1 and SS 2), and WRG (best seen in color; edge directions and punctuation omitted for readability). Vertices are word tokens and their indices, edges are neighbor (\textit{ngbh}) and/or dependency relations.  
Dashed lines represent edges to \textbf{B}reak, the green curved line represents an edge to \textbf{C}opy, the open circle node for \texttt{and-6} is for \textbf{D}rop, and all other parts of the graph get \textbf{A}ccept. At bottom left is a fragment of the corresponding Edge Triple Set.
}
    \label{fig:wcg}
\end{figure}
Figure~\ref{fig:wcg} shows an example sentence and a simplified version of its WRG (edge directions are not shown, for readability). Vertices are labeled with word-index pairs in red font, and edges are labeled as 
\textit{ngbh} for neighboring words, or with the tags corresponding to their dependency relations, such as \textit{nsubj} between \textit{Sokuhi-1} and \textit{ordained-13}. An edge can have both types of relation, e.g. neighbor and dependency for \textit{was-12} and \textit{ordained-13}. 
The graph is stored as an Edge Triple Set, a set of triples with (source node, target node, label) representing each pair of words connected by an edge, as shown in Figure~\ref{fig:wcg}, bottom left. 
Given a sentence and its WRG, our goal is to decompose the graph into $n$ connected components (CC) where 
each CC is later rewritten as an output simple sentence. 
To perform the graph decomposition, decisions are made on every edge triple.
We define four \textit{edit} types: 
\vspace{-.1in}
\begin{itemize}
\setlength\itemsep{-0.5em}
\item \texttt{\textbf{A}ccept}: retain the triple in the output 
\item \texttt{\textbf{B}reak}: break the edge between a pair of words
\item \texttt{\textbf{C}opy}: copy a target word into a CC 
\item \texttt{\textbf{D}rop}: delete the word from the output CCs
\end{itemize}
\vspace{-.1in}
A training example consists of an input sentence, and one or more output sentences. If the input sentence is complex, the ground truth output consists of multiple simple sentences. The next section 
presents the \ours pipeline. Two initial modules construct the WRG graphs for each input sentence, and the \ours labels for the Edge Triple Sets based on the ground truth output. A neural model learns to assign \ours labels to input WRG graphs, and a final graph segmenter generates simple sentences from the labeled WRG graphs. Details about the neural model are in the subsequent section.


\section{System Overview}


\begin{figure*}
 \centering
\includegraphics[scale=0.35]{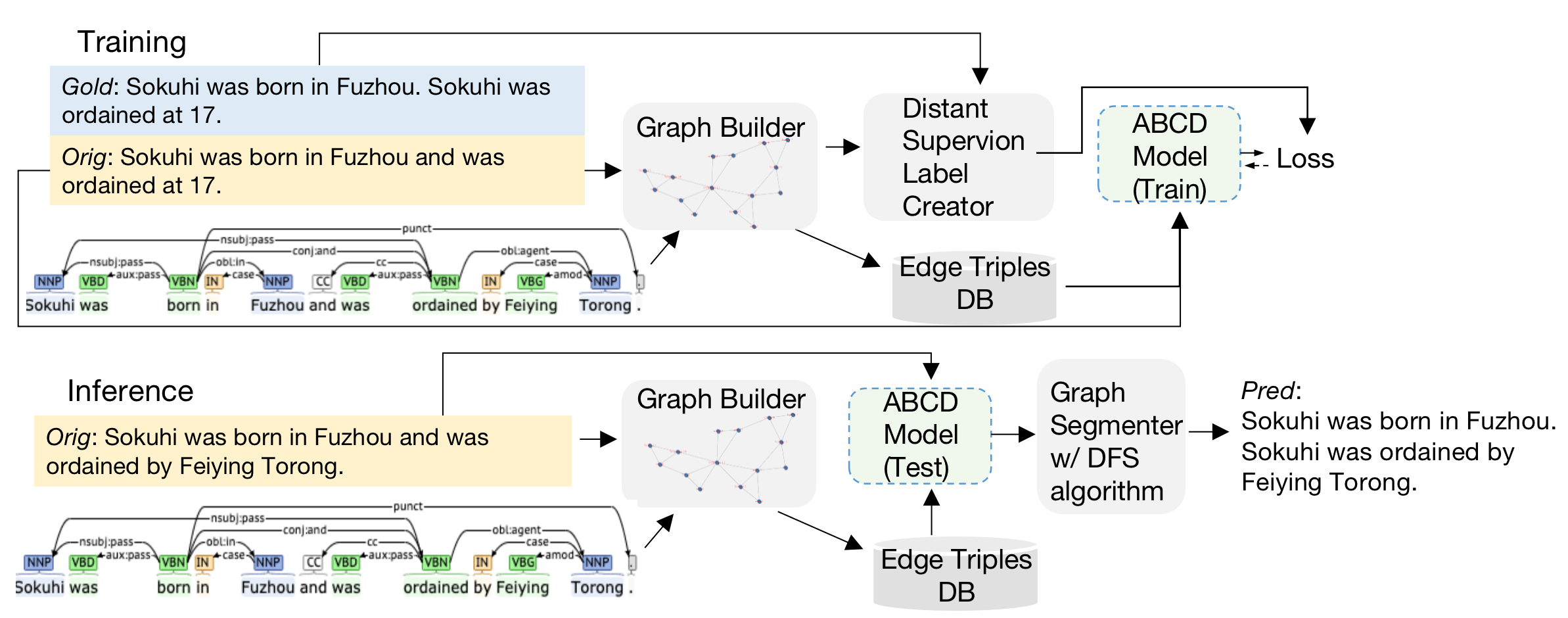}
\caption{\small \ours system overview during training (top) and inference (bottom).}
    \label{fig:system}
\end{figure*}

The full processing pipeline 
consists of 
five major components, as shown in Figure~\ref{fig:system}. Three pre-processing modules handle the WRG graph construction, conversion of graph triples to vectors, and creation of distant supervision labels for the graph. The fourth component is the \ours neural model that learns to label a WRG graph, which is described in section~\ref{sec:model}. The last part of the pipeline is a post-processing module to segment WRG graphs based on the labels learned by the \ours model, and to map each graph segment to a simple sentence.   

\paragraph{Graph Constructor}
The first module in the system is a Graph Constructor that converts an input sentence and its dependency parse into a collection of vertices and edges. It is used during training and inference. 
It first extracts words and their indices from the input sentences of the training examples for the vertices of each WRG graph. 
A directed edge and \textbf{ngbh} label
is assigned to all pairs of adjacent words. A directed edge and label is also assigned to every governing and dependent word pair 
(cf. Figure~\ref{fig:wcg}).

\paragraph{Edge Triples DB}

The Edge Triples DB, which is used during training and inference, creates vector representations for the input Edge Triples Sets for each training instance, using latent representations learned by an encoder component of the \ours model. Using the word indices, a function maps the source and target words from every triple into its hidden representation learned by the encoder, and the triple's edge label is converted into a one-hot encoding with dimension $d$.  For an edge triples set with $m$ triples, the source and target word hidden states are each stacked into an $m \times h$ matrix, and the one-hot vectors for edge labels are stacked into an $m \times d$ matrix. These three source, target, edge matrices that represent an Edge Triple Set are then fed into an attention layer, as discussed in section~\ref{sec:model}.

\paragraph{Distant Supervision Label Creator}
The expected supervision for our task is the choice of edit type 
for each triple, where the ground truth consists of 
pairs of an input sentence, and one or more output simple sentences.
We use distant supervision where we automatically create edit labels 
for each triple based on
the alignment between the original input sentence and the set of output simple sentences. 
In the Distant Supervision Label Creator module, 
for every triple, 
we check the following conditions: if the 
edge is a "neighbor" relation, and both source and target words are in the same output simple sentence, 
we mark this pair with edit type \texttt{A}; if the source and target words of a triple occur in different 
output simple sentences, the corresponding edit is \texttt{B}; if the source and target are in the same output simple sentence, 
and the only edge is a dependency label (meaning that they are not adjacent in the original sentence), we mark this pair as \texttt{C}; finally, if a word 
is not in any output simple sentence, we mark the corresponding type as \texttt{D}. 


\paragraph{Graph Segmenter}
This module 
segments the graph into connected components using predicted edits, and 
generates the output sentences, as part of the inference pipeline. 
There are four stages consisting of:
graph segmentation, 
traversal, subject copying, and output rearranging. In the graph segmentation stage, the module first performs actions on every triple 
per the predicted edit: if the edit is \texttt{A}, no action is taken; if the edit is \texttt{B}, the edge between the pair of words is dropped; given \texttt{C}, the edge is dropped, and 
the edge triple is stored 
in a temporary list for later retrieval; if the edit is \textsc{d}, the target word is dropped from the output graphs. 
After carrying out the predicted edits, we run a graph traversal algorithm on modified edge triples to find all CCs, 
using a modified version of the Depth-First-Search algorithm with linear time proposed in~\cite{tarjan1972depth, nuutila1994finding}. 
For each CC, the vertices are kept and the edges are dropped.
Then we enter the subject copying stage: for each source, target pair in the temporary list mentioned earlier, we copy the word to the CC containing the target.
Finally for every CC, we 
arrange all words in their linear order by indices, and output a simple sentence.     


\begin{figure}[h]
 \centering
\includegraphics[width=\columnwidth]{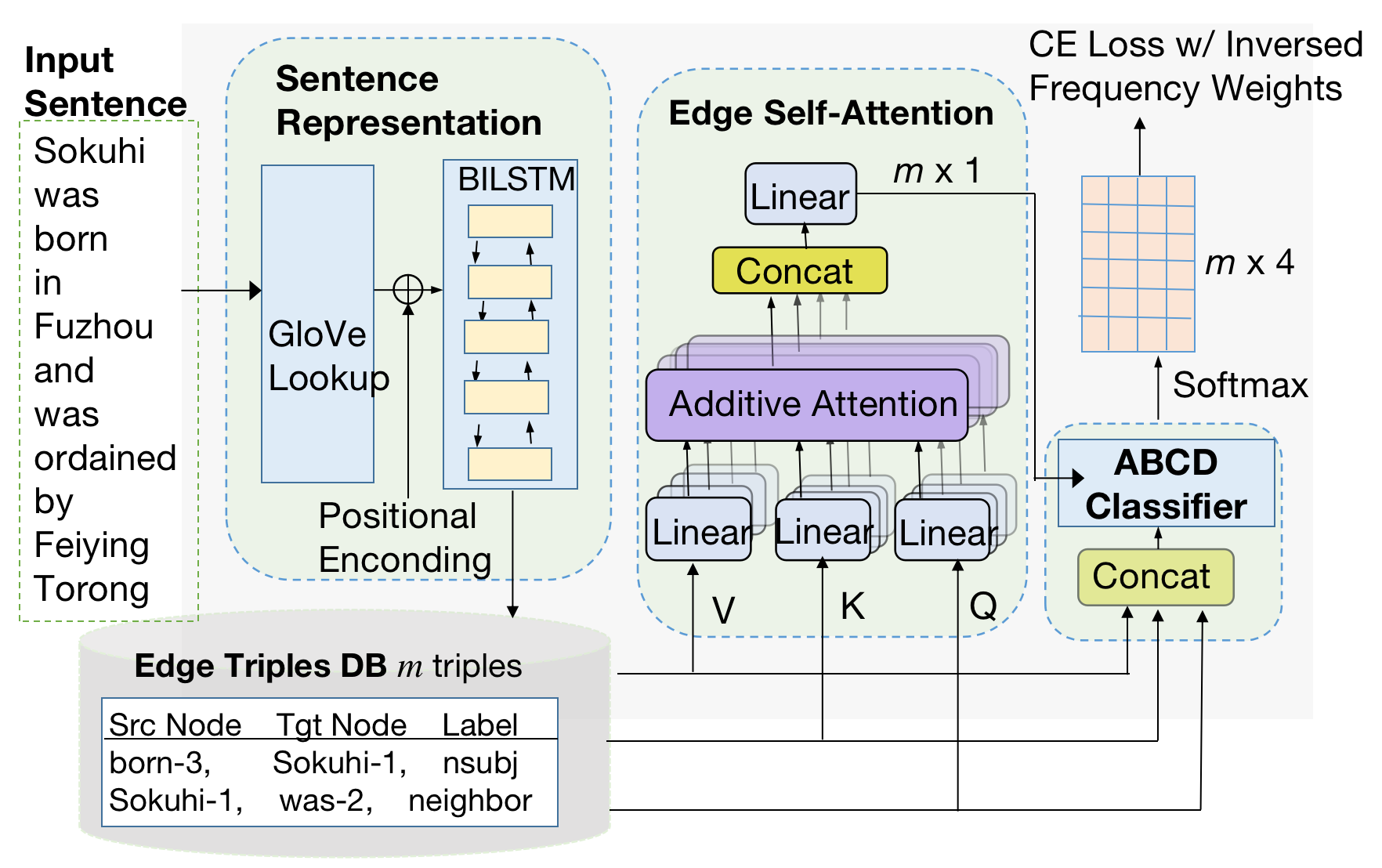}
\vspace{-.15in}
\caption{\small Architecture for \ours model.}
    \label{fig:neural}
\end{figure}

\section{Neural Model}
\label{sec:model}

The \ours model consists of three neural modules 
depicted in Figure~\ref{fig:neural}: 
a sentence encoder to learn a hidden representation for the 
input sentence, a self-attention layer to generate attention scores on every edge label, and a classifier that 
generates a predicted distribution over the four edit types, based on the word's hidden representation, the edge label representation, and the attention scores.


\subsection{Sentence Representation}
The sentence representation module 
has two components: a word embedding look up layer based on GloVe~\cite{pennington2014glove}, and a bidirectional LSTM~\cite{hochreiter1997long} (see Figure~\ref{fig:neural}). Given an input sentence length \textit{l}, and the hidden state dimension $M$, the output from this module is $l \times M$. For a word with index \textit{i} in the input sentence, we 
generate its hidden representation $h_i$ such that it combines the hidden states from forward and backward LSTMs, with $h_i \in \mathbb{R}^{M}$. 
A positional encoding function is added to the word embeddings. 
We found this particularly helpful in our task, 
presumably because 
the same word type at different positions might have different relations with other words, captured by distinct learned representations. Our experiments compare biLSTM training from scratch to use of  BERT~\cite{devlin-etal-2019-bert}, to see if pre-trained representations are helpful.  

To utilize the 
learned 
word representations 
in the context of the relational information captured in the WRG graph,
we send the sentence representation to the Edge Triple DB and extract representations $h_i$ and $h_j$ for the source 
and target words, based on indices $i$ and $j$. A one-hot vector 
with dimensionality $N$
encodes relations between pairs of 
source and target 
words; 
each edge triple is thus converted into three vectors: $h_{src}$, $h_{tgt}$, $d_{rel}$. We take position-wise summation over all one hot vectors if there is more than one label on an edge. 


\subsection{Edge Self-Attention}

Attention has been 
useful for many NLP tasks. In our model, we adapt the multi-head self attention mechanism~\cite{vaswani2017attention} to learn 
importance
weights on types of edit operations, as shown in the middle green block in Figure~\ref{fig:neural}. Given $m$ edge triples, we first stack all source vectors $h_{src}$ into a matrix $H_{src}$, and operate the same way on $h_{tgt}$ and $d_{rel}$ to obtain $H_{tgt}$ and $D_{rel}$, such that $H_{src}, H_{tgt}  \in \mathbb{R}^{m \times M}$, and $D_{rel} \in \mathbb{R}^{m \times N}$. These three matrices are the input to self-attention. 
For every head of the multi-head attention, we first obtain a feature representation with the three parameters $V, K, Q$ mapping to sources, targets and relations, respectively, then compute a co-efficient 
$e$ with a learnable parameter $W^e$
as follows:
\begin{equation}
 \begin{aligned}
      e = LeakyRelu(W^{e}(VH_{src}; KH_{tgt}; QD_{rel}))
      \end{aligned}
\end{equation}
where $e \in \mathbb{R}^{m \times 1}$. 
Then we compute the attention scores by taking a softmax over $e$: 
\begin{equation}
      head = softmax(e)
      \vspace{-.5pc}
\end{equation}
Finally, we concatenate all head attentions together, and pass 
them
through a linear layer to learn the relations between heads, and generate the final attention scores: 
\begin{equation}
      \alpha = W(concat((head_1, head_2, \ldots))  
      \vspace{-.5pc}
\end{equation}
 $\alpha \in \mathbb{R}^{m \times 1}$. 
The attention scores are sent to the next module to help the classifier make its decision.

\subsection{Edit Classification}

The last component of our neural model is a classifier, as shown 
at the right
of Figure~\ref{fig:neural}. To aggregate the feature representation from the previous layer, we first concatenate the three matrices $H_{src}$, $H_{tgt}, D_{rel}$ into one representation, and multiply the attention scores as follows:
\vspace{-.1in}
\begin{equation}
      H^{\prime} = \alpha(H_{src}; H_{tgt}, D_{rel})   
\end{equation}
An MLP layer then takes $H^{\prime}$ as its input and generates the output distribution over the four edit types 
for each edge triple:
\begin{equation}
      Out_{M} = Softmax(MLP(H^{\prime}))  
\end{equation}
where $Out_M \in \mathbb{R}^{m \times 4}$. 

As an alternative to MLP,
we also investigated a bilinear classifier, which has proved efficient in capturing fine-grained differences in features for classification task~\cite{DBLP:conf/iclr/DozatM17}. The bilinear layer first takes $H_{src}$ and $H_{tgt}$ as input and generates transposed bilinear features : 
\vspace{-.1in}
\begin{equation}
      output^{bi} = H_{src}^{\intercal}W^AH_{tgt} + b 
\end{equation}
where $W^A$, $b$ are learnable parameters. Then we sum the bilinear features with the MLP decisions and apply softmax on the result to get the final distribution over the four edit labels: 
\begin{equation}
      Out_{B} = Softmax(output^{bi} + MLP(H^{\prime}))   
\end{equation}
where $Out_B \in \mathbb{R}^{m \times 4}$.  We use cross entropy loss between predicted edit types and gold edit types created from distant supervision
(see above). 

\subsection{Training}
\begin{table}[t]
\small 
    \centering
    \begin{tabular}{l|rrrr} \thickhline 
       Dataset  & A & B & C & D  \\  \hline
       MinWiki & 85.23\% & 4.58\% & 3.60\% & 6.57\% \\ 
        DeSSE & 74.77\% & 2.39\% & 5.62\% & 17.21\% \\ 
            \hline 
        MinWiki & 0.0167 & 0.3533& 0.4164 & 0.2135 \\ 
        DeSSE & 0.0200 & 0.6266 & 0.2658 & 0.0876 \\ 
    \thickhline 
    \end{tabular}
    \caption{Distributions (Top) and inverse class weights (Bottom) for the four edit labels on both MinWiki and DeSSE datasets.}
    \label{tab:distant_supervision}
\end{table}
The class balance for our task is highly skewed:
the frequency of class \texttt{A} is much higher than the other three classes, as shown in the top portion of Table~\ref{tab:distant_supervision}. To 
mitigate the impact 
on training, we adopt the inverse class weighting for cross entropy loss introduced in~\cite{huang2016learning}. With this weighting, loss is weighted heavily towards rare classes, 
which
forces the model to learn more about the rare cases. Table~\ref{tab:distant_supervision} shows the weights for four edit labels on both datasets. 
On MinWiki, \texttt{A} occurs the most and has the lowest weights as 0.0167, a sharp contrast to \texttt{B,C,D}. On DeSSE, both \texttt{A} and \texttt{D} occur frequently while \texttt{B} and \texttt{C} have lower frequency with higher weights, at 0.6266 and 0.2658. 
DeSSE has fewer \texttt{B}, and more \texttt{C} and \texttt{D} than MinWiki. From this perspective, MinWiki is ``simpler'' than DeSSE because there are fewer edits on rewriting the sentences. This might be due to the different distributions of linguistic phenomena in the two datasets (see Table~\ref{tab:lings}). In the next section, we will show that \ours shows stronger improvements on complicated edits. Training details are in the appendix. 


\section{Experiments}
\begin{table*}[ht!]
    \small
    \centering
    \begin{tabular}{l|rrrr|rrrr} \thickhline
    \multirow{3}{*}{Category} & \multicolumn{4}{c|}{MinWiki} &  \multicolumn{4}{c}{DeSSE} \\ 
    & \multicolumn{2}{c}{biLSTM} &\multicolumn{2}{c|}{BERT} & \multicolumn{2}{c}{biLSTM} &\multicolumn{2}{c}{BERT} \\  
  & mlp & bilinear & mlp & bilinear & mlp & bilinear & mlp & bilinear  \\ \hline
  A  & 0.98 & 0.98 &0.93& 0.86 & 0.91& 0.88 &0.88 &0.87 \\
    B & 0.48 &0.48& 0.41 & 0.36 &0.34& 0.42 &0.31 &0.28  \\
    C &0.99& 0.99 & 0.95&0.98 & 0.89& 0.78 & 0.89 &0.55 \\ 
    D &0.80 & 0.84 &0.39& 0.75 & 0.49& 0.54 & 0.45 &0.45\\ \hline 
    All & 0.78 & \textbf{0.82} & 0.72 & 0.74 & 0.66 & \textbf{0.67} & 0.63 & 0.57\\ 
    \thickhline
    \end{tabular}
    \vspace{-0.7pc}
    \caption{\small Performance (F1) of our model and its variants on MinWiki (N=1075) and DeSSE (N=790).  }
    \label{tab:abcd}
\end{table*}
We carry out
two intrinsic evaluations of 
\ours performance on MinWiki and DeSSE. Section~\ref{sec:edit} presents an intrinsic evaluation of \ours variants 
on edit prediction, with error analysis and ablation studies. Section~\ref{sec:prop} 
compares the best \ours model with several baselines on the quality of output propositions. We 
discuss evaluation metrics in section~\ref{sec:analysis}. Results show that \ours models show consistently good performance compared to other baseline models on both datasets.

\begin{table}[t]
\small 
    \centering
    \begin{tabular}{l|l|rrrr} \thickhline 
        \multirow{2}{*}{Data} &  \multirow{2}{*}{Gold} &  \multicolumn{4}{c}{Predicted}   \\
         &  & A & B & C & D \\   \hline 
         \multirow{2}{*}{Minwiki} & B & 36.10 & 48.33 & 5.59& 9.98 \\ 
      & D & 14.01 & 0.14 & 0.46 & 85.38  \\ 
      \multirow{2}{*}{DeSSE} & B & 27.42 & 46.62 & 15.18& 10.76 \\ 
      & D & 42.63 & 3.44 & 5.08 & 48.84  \\ \thickhline 
    \end{tabular}
    \vspace{-0.7pc}
    \caption{\small Percentage (\%) of count of predicted labels where gold labels are \texttt{B} and \texttt{D} from \ours\textsubscript{biLSTM+mlp}. }
    \label{tab:decision_b_d}
\end{table}

\begin{table*}[t]
    \small
    \centering
    \begin{tabular}{ll|rrrr|rrrr} \thickhline
      \multirow{3}{*}{Group} & \multirow{3}{*}{Model} & \multicolumn{4}{c|}{MinWiki} & \multicolumn{4}{c}{DeSSE}  \\ 
      & & \#T & Match  & BLEU4  & BERTSc & \#T & Match & BLEU4 & BERTSc  \\ 
      & &/SS & \#SS(\%) & & &/SS & \#SS(\%) & &   \\ \hline 
      \multirow{4}{*}{Parsing} & DisSim &  8.50 & 68.46 & 64.20 & 94.42 & 9.59 & 40.00 & 37.89 & 89.54   \\ 
            & DCP$_{vp}$ & 14.82 & 45.49 & 28.80 &64.50 & 15.99 & 42.40 & 47.25 & 60.18\\ 
            & DCP$_{sbar}$ & 19.07 & 17.49  & 19.35 & 49.07 & 17.24 & 44.81 & 48.02 & 59.89  \\
            & DCP$_{recur}$ &16.30 &67.90 & 31.78 &58.08 & 14.16 & \textbf{55.63} & 34.44 & 61.37 \\ \hline
         \multirow{1}{*}{Encoder-decoder}  & COPY & 9.37 & \textbf{79.26} & \textbf{80.96} & \textbf{95.96} & 18.13 & 36.20 & 45.91 & 88.71 \\
            \hline
         \multirow{2}{*}{\ours biLSTM} &  mlp & 9.37 & 78.61 & 75.80 & 92.91 & 8.85 & 53.29 & \textbf{53.42} & 90.23\\ 
            & bilin & 9.53 & 76.72 & 76.38 & 90.28 &8.10 &52.66 & 41.57 & \textbf{94.78} \\ 
            \hline
          \thickhline
    \end{tabular}
    \vspace{-0.8pc}
    \caption{\small Performance of baselines and our models on Minwiki test set (N=1075, \#T/SS = 10.03), and DeSSE test set (N=790, \#T/SS =9.07). We report numbers of token per propositions (\#T/SS), number of input sentences that have match number of output between prediction and ground truth in percentage (Match \#SS\%), BLEU with four-gram and BERTScore.}
    \label{tab:baselines}
\end{table*}

\subsection{Intrinsic Evaluation on Edit Prediction}
\label{sec:edit}

We report F1 scores on all four edit types from \ours and its model variants. We compare two classifiers as mentioned in previous sections and investigate the difference between using biLSTM and BERT with fine-tuning, to see if pre-trained knowledge is useful for the task. 

Table~\ref{tab:abcd} presents results on MinWiki and DeSSE from the four model settings. 
All models perform better on MinWiki than DeSSE, and biLSTM+bilinear shows the best performance on both, with F1 scores of 0.82 and 0.67 on MinWiki and DeSSE respectively. Presumably this reflects the greater linguistic diversity of DeSSE shown in Table~\ref{tab:lings}.
The lower performance from BERT variants indicates the pre-trained knowledge 
is not helpful.
Among the four edit types, all models have high F1 scores on \texttt{A} across datasets, high F1 on \texttt{C} for MinWiki, but not on DeSSE. \texttt{B} and \texttt{D} show lower scores, and all four models report lower F1 on \texttt{B} than \texttt{D} on both datasets. 

To examine the significant drop on \texttt{B} and \texttt{D} from MinWiki to DeSSE, Table~\ref{tab:decision_b_d} presents error 
analysis on pairs of gold labels and predictions 
for \texttt{B} and \texttt{D}, using predictions from biLSTM+mlp. The model does poorly 
on \texttt{B} in both datasets, compared with 
predictions of 36.1\% for \texttt{A} on MinWiki, on on DeSSE, 27.42\% for \texttt{A} and 15.18\% for \texttt{C}. The model has high agreement on \texttt{D} from MinWiki, but predicts 42.63\% \texttt{A} on DeSSE. We suspect that improved feature representation could raise performance; that is, pairs of words and their relations might be a weak supervision signal for \texttt{B} and \texttt{D}. 

We conducted an ablation study on the inverse class weights mentioned in section 6 on MinWiki. After removing the weights, the model fails to learn other classes and only predicts \texttt{A} due to the highly imbalanced label distributions, which demonstrates the 
benefit of weighting the loss function.
We also ablate positional encoding 
which leads to F1 scores of 0.90 for \texttt{A}, 0.51 for \texttt{C}, and 0 for both \texttt{B} and \texttt{D}, indicating the importance of positional encoding. 

\subsection{Intrinsic Evaluation of Output Sentences}
\label{sec:prop}

\begin{table}[ht]
    \small
    \centering
    \begin{tabular}{l|rrr|rrr} \thickhline
    \multirow{2}{*}{} & \multicolumn{3}{|c|}{MinWiki (N=1075)} & \multicolumn{3}{c}{DeSSE\textsubscript{pos}(N=521)}  \\ 
       & M & BL4 & BS & M & BL4 & BS  \\  \hline 
            Copy & 852 & 88.81 & 97.16   & 20 & 92.48 & 98.66 \\
            \hline
            mlp  & 845  & 89.59 & 97.21 & 247 & 78.49 & 95.73 \\
            bilin & 825  & 92.00 & 96.94  & 251 & 74.25 & 98.21 \\ 
            \hline
          \thickhline
    \end{tabular}
    \vspace{-0.7pc}
    \caption{\small Performance of Copy\textsubscript{512} and our \ours biLSTM models on all positive samples from MinWiki and DeSSE test set. Columns show the raw count of complex sentences where prediction has correct number of outputs (M), BL4 and BS. }
    \label{tab:closer}
\end{table}

For baselines, we use Copy\textsubscript{512} and DisSim, which both report performance on Wikisplit in previous work. 
We also include DCP, 
which relies on three rules applied to token-aligned dependency and constituency parses: DCP\textsubscript{vp} extracts clauses with tensed verb phrases; DCP\textsubscript{sbar} extracts \texttt{SBAR} subtrees from constituency trees; DCP\textsubscript{recur}
recursively applies the preceding rules.

For evaluation, we use BLEU with four-grams (BL4) \cite{papineni2002bleu} and BERTScore (BS)~\cite{zhang2019bertscore}. We also include descriptive measures specific to our task. To indicate whether a model retains roughly the same number of words as the source sentence in the target output,
we report average number of tokens per simple sentence (\#T/SS). 
To capture the correspondence between the number of target simple sentences in the ground truth and model predictions,
we use percentage of samples where the model predicts the correct number of simple sentences (Match \#SS). 
BL4 captures the 4-gram alignments between candidate and reference word strings, 
but fails to assess similarity of latent  meaning. BS applies token-level matching through contextualized word embeddings,  therefore evaluates candidates on their word meanings. For each example, we first align each
simple sentence in the ground truth with a prediction, compute the pairwise BL4 and BS scores, and take the average as the score for the example. 
A predicted output sentence with no correspondent in the ground truth, or a ground truth sentence with no correspondent in the predicted, will add 0 to the numerator and 1 to the denominator of this average.

Table~\ref{tab:baselines} presents results from the baselines and our \ours best variant, biLSTM with two classifiers. None of the models 
surpasses all others
on both datasets. 
All models show lower performance on DeSSE than MinWiki, 
again an indication that DeSSE is more challenging.
On MinWiki, 
\ours is competitive with Copy\textsubscript{512},
the best performing model, with a narrow gap on Match\#SS (0.65\%) and BLEU4 (4.58). On DeSSE, \ours BL4 and BS surpass all baselines. 
\ours performance is 2.34\% less than DCP\textsubscript{recur} on Match \#SS, but biLSTM+mlp
output sentences have an average length of 8.85, which is closer to the gold average length of 9.07, in contrast to much longer output from DCP\textsubscript{recur} of 14.16.
To summarize, \ours achieves competitive results on both datasets.

\subsection{Error Analysis}
\label{sec:analysis}




While Table~\ref{tab:decision_b_d} presents error analysis on 
predictions of \texttt{B} that lead to an incorrect number of outputs, here we 
examine
test sentences from both datasets where the prediction and ground truth have the same number of outputs. 
Table~\ref{tab:closer} shows the total number of examples for MinWiki (1,075) and for the positive examples in DeSSE (DeSSE$_{pos}$, 521). The M columns for each dataset give the number of examples where the number of targets in the ground truth matches the number of targets predicted by the model.
On MinWiki, 
\ours has marginally better BL4 and BS scores than Copy\textsubscript{512}, but 
Copy\textsubscript{512} has 7 more cases with the correct number of outputs. 
For DeSSE, we restrict attention to the positive examples (MinWiki has no negative examples), because Copy\textsubscript{512} and \ours perform equally well on the negative examples.
By the BL4 and BS scores on DeSSE$_{pos}$, Copy\textsubscript{512} appears to perform much better than \ours, but these scores are on 20 out of 521 examples (3.8\%). Although \ours's scores are lower, it produces the correct number of output sentences in 47.4\% of cases for the mlp, and 48.1\% for the bilin.

\begin{figure*}
\small 
\centering
\begin{tabular}{l|l} \thickhline
     Orig & He did not do anything wrong, yet he was targeted and his family was murdered. \\ 
     Human & He did not do anything wrong. $\vert \vert$ He was targeted. $\vert \vert$ His family was murdered. \\ 
     Copy & He did not do anything wrong, he was targeted and his family was murdered. \\ 
     \ours & He did not do anything wrong.$\vert \vert$ he was targeted. $\vert \vert$ his family was murdered.  \\ \hline 
     Orig & I guess I always knew it was \textbf{genetics} but I didnt know why our features are the way that they are. \\ 
     Human & I guess I always knew it was genetics. $\vert \vert$ I didnt know why our features are the way that they are.\\ 
     Copy & I guess I always knew it was \textcolor{red}{interesting}.$\vert \vert$ I didnt know why our features are the way that they are. \\ 
     \ours & I guess I always knew it was \textcolor{blue}{genetics}.$\vert \vert$ I didnt know why our features are the way that they are. \\ \hline
     Orig & Our main mission, which has been the main mission since 9/11 is to eliminate terrorism wherever it may exist.  \\ 
     Human & Our main mission, which has been the main mission since 9/11.$\vert \vert$ It is to eliminate terrorism wherever it may exist.\\ 
     Copy &  \textit{same as Orig}  \\ 
     \ours & Our main mission has been the main mission. $\vert \vert$ \textit{mission} is to eliminate terrorism wherever it may exist. \\ \thickhline
     
     \thickhline
\end{tabular}
\caption{Three input complex sentences (Orig) from DeSSE, with the annotated rewriting (Human), and the predicted propositions from Copy and \ours. }
\label{fig:output_example}
\end{figure*}

Figure~\ref{fig:output_example} shows three complex sentences from DeSSE with the annotated rewriting, and predicted propositions from Copy$_{512}$ and \ours$_{mlp}$. Copy$_{512}$ correctly decomposes only one of the examples and copies the original input on the other two samples. On the one example where Copy produces two simple sentences, it alters the sentence meaning by replacing the word ``genetics" with the word ``interesting". This exposes a drawback of encoder-decoder architectures on the proposition identification task, that is, the decoder can introduce words that are not in the input sentence, 
therefore failing to preserve the original meaning. 
In contrast, \ours shows good performance on all three sentences by producing the same number of simple sentences as in the annotated rewriting. Especially for the third sentence, which contains an embedded clause, \textit{``which has been the main mission since 9/11"}, the first proposition written by the annotator is not grammatically correct, and the subject of the second proposition is a pronoun \textit{it}, referring to the semantic subject \textit{Our main mission}. Nonetheless, \ours generates two propositions, both of which are grammatically correct and meaning preserving.  


\section{Discussion}

In this section, we 
discuss limitations of ABCD to 
guide future work. The first limitation is 
the low performance of ABCD on \texttt{B}. We observe that in DeSSE, some annotators did not break the sentences appropriately. We randomly selected 50 samples, and found 13 out of 50 (26\%) 
examples where annotators add breaks to rewrite NPs and infinitives as clauses. This introduces noise into the data. Another reason of lower performance on \texttt{B} might be attributed to the current design of \ours that neglects sequential relations among all words. Among all edge triples where it fails to assign B,  67\% and 27.42\% are with \texttt{ngbh} relations on MinWiki and DeSSE,
respectively. 
Two possibilities for improving performance to investigate are enhancements to the information in the WRG graph, and re-formulating the problem 
into sequence labeling of triples.

The second limitation pertains mainly to DeSSE. In the training data, 34.7\% of sentences have OOV words. 
 For example, we noticed that annotators sometimes introduced personal pronouns (e.g.\textit{he/she/they}) in their rewrites of VP-conjunction, 
instead of copying the subjects, or they substituted a demonstrative pronoun (e.g.\textit{this/these}) for clausal arguments.
 This could be addressed by 
 expanding the edit types to include the ability to \textsc{insert} words
 from a restricted insertion vocabulary. Nevertheless, our model has a small performance gap with Copy\textsubscript{512} on MinWiki, and outperforms 
 the baselines on DeSSE. 

A third issue is whether ABCD would generalize to other languages. We expect ABCD would perform well on European languages with existing dependency and constituency parsers, and with an annotated dataset.

\section{Conclusion}

We presented a new task to decompose complex sentences into simple ones, along with DeSSE, a new dataset designed for this task. We proposed the neural ABCD model to predict four edits operations 
on sentence graphs, 
as part of
a larger pipeline from 
our graph-edit problem formulation.
ABCD performance 
comes close to or outperforms
the parsing-based and encoder-decoder baselines. 
Our work selectively integrates modules to capitalize on the linguistic precision of parsing-based methods, and the expressiveness of graphs for encoding different aspects of linguistic structure, while still capitalizing on the 
power 
of neural networks for representation learning.   


\bibliographystyle{acl_natbib}
\bibliography{acl2021}

\clearpage
\appendix
\newpage
\section{Annotation Instruction in DeSSE}
\label{sec:appendixB}

\begin{figure}[ht]
\small 
\centering
\includegraphics[width=\columnwidth]{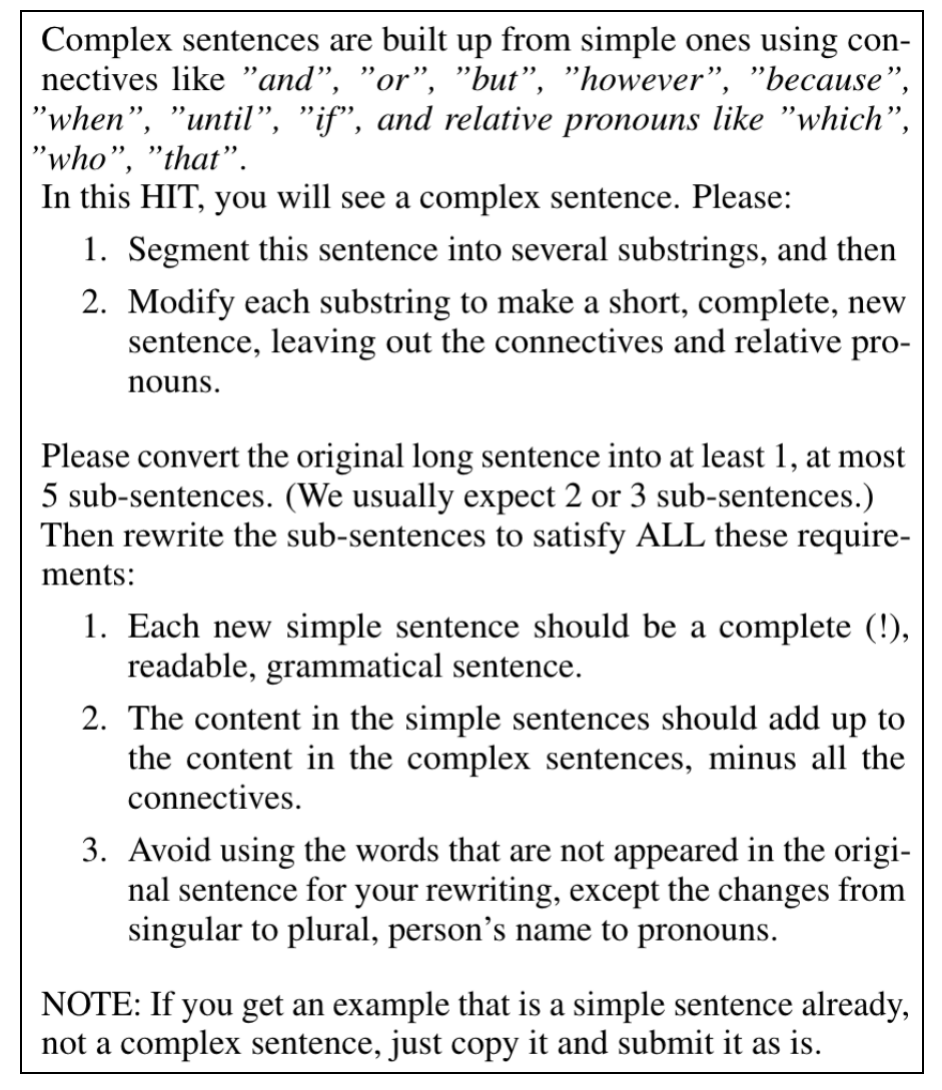} 
\vspace{-1.5pc}
    \caption{Instruction for DeSSE annotation}
    \label{fig:inst}
\end{figure}

Here we present the instructions for annotators, as shown by Figure~\ref{fig:inst}.

\begin{figure}[ht]
\centering
\includegraphics[width=\columnwidth]{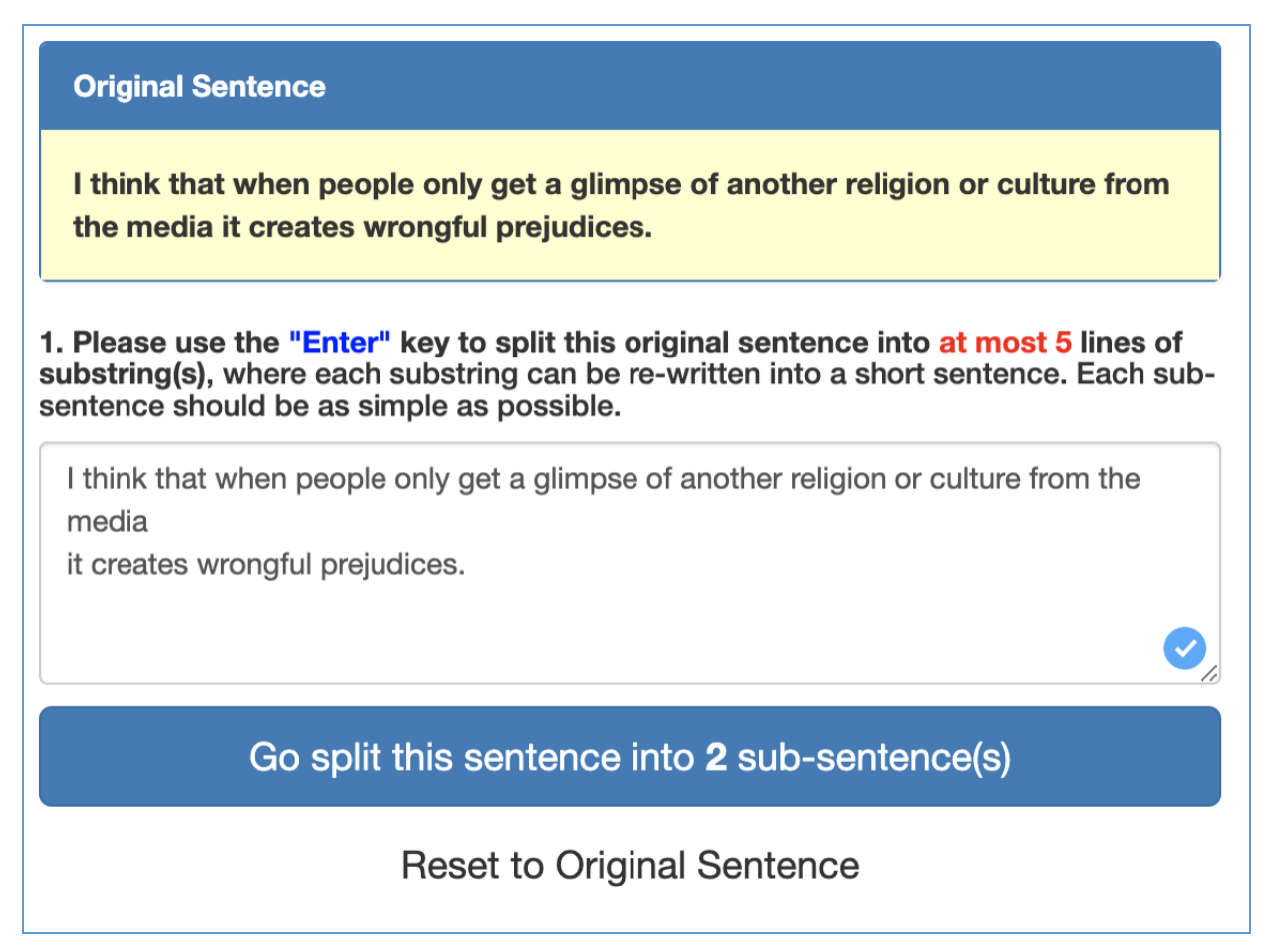} 
\vspace{-2pc}
\caption{Interface of splitting the sentence} 
\label{fig:ann_interface1}
\end{figure}

\begin{figure}[ht]
\centering
\includegraphics[width=\columnwidth]{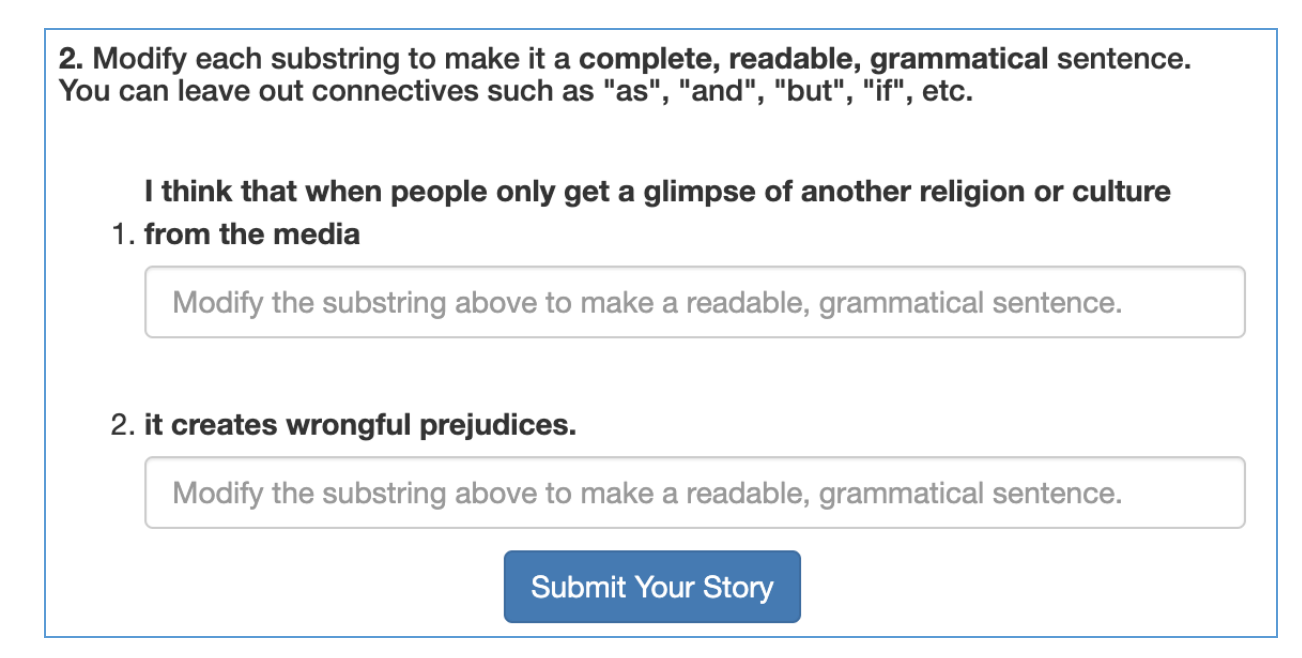} 
\caption{Interface of rewriting the segments from Figure~\ref{fig:ann_interface1} into complete sentences} 
\label{fig:ann_interface2}
\end{figure}

The instructions illustrate the two phases of annotation. The annotator first chooses whether to add one or more split points to an input sentence, where the word after a split point represents the first word of a new segment. Once an annotator has identified the split points (first page of the AMT interface, shown as Figure~\ref{fig:ann_interface1}), a second page of the interface appears. Figure~\ref{fig:ann_interface2} shows the second view when annotators rewrite the segments. Every span of words defined by split points (or the original sentence if no split points), appears in its own text entry box for the annotator to rewrite. Annotators cannot submit if they remove all the words from a text entry box. They are instructed to rewrite each text span as a complete sentence, and to leave out the discourse connectives. 

 Several kinds of auto-checking and warnings are applied in the interface to ensure quality.  If a rewrite contains a discourse connective, a warning box pops up asking if they should drop the discourse connective before submitting it. A warning box will show up if annotators use vocabulary outside the original sentence. To prevent annotators from failing to rewrite, we monitored the output, checking for cases where they submitted the text spans with no rewriting. Annotators were prohibited to submit if the interface detected an empty rewrite box or the total lengths of the rewrites are too short compared to the source sentence. We warned annotators by email that if they failed to produce complete sentences in the rewrite boxes, they would be blocked. Some annotators were blocked, but most responded positively to the warnings. 
 
\section{Quality control in DeSSE}
\label{sec:appendixC}

To test the clarity of instruction and interface, the initial 500 sentences were used for evaluating the task quality, each labeled by three turkers (73 turkers overall), using three measures of consistency, all in [0,1]. Average pairwise boundary similarity~\cite{fournier2013evaluating}, a very conservative measure of whether annotators produce the same number of segments with boundaries at nearly the same locations, was 0.55.  Percent agreement on number of output substrings was 0.80. On annotations with the same number of segments, we measured the average Jaccard score (ratio of set intersection to set union) of words in segments from different annotators, which was 0.88, and words from rephrasings, which was 0.73. With all metrics close to 1, and boundary similarity above 0.5, we concluded quality was already high. During the actual data collection, quality was higher because we monitored quality on a daily basis and communicated with turkers who had questions. 

\section{Experiment Settings}
We trained our model on a Linux machine with four Nvidia RTX 2080 Ti GPUs. We conducted grid search for the hyper-parameters, with learning rage in the range of [1e-2, 1e-5] (step size 0.0005), weight decay between [$0.90$, $0.99$], hidden size [200, 800] (step size 200). Final parameters are set with Adam optimizer and learning rate at $1e-4$, weight decay $0.99$, embedding dropout at $0.2$, maximum epoch as 100 with early stop. We use GloVe 100 dimension vectors, hidden size of network as 800. We set the number of heads in self-attention as 4, corresponding to the four edit types. With batch size 64, it takes about 6 hours to train MinWiki and 4 hours for DeSSE.  For BERT fine-tuning, we use $1e-4$ learning rate, weight decay at $0.99$.

\end{document}